# BERTCaps: BERT Capsule for Persian Multi-Domain Sentiment Analysis

Mohammadali **Memari**, Soghra Mikaeyl **Nejad**, Amir Parsa **Rabiei**, Mehrshad **Eisaei** and Saba **Hesaraki**

ARTICLE INFO

ABSTRACT

Multi-domain sentiment analysis involves estimating the polarity of an unstructured text by exploiting domain-specific information. One of the main issues common to the approaches discussed in the literature is their poor applicability to domains that differ from those used to construct opinion models. This paper aims to present a new method for Persian multi-domain SA analysis using deep learning approaches. The proposed BERT-Capsules (BERTCaps) approach consists of a combination of BERT and Capsule models. In this approach, BERT was used for Instance representation, and Capsule Structure was used to learn the extracted graphs. Digikala's dataset, including ten domains with both positive and negative polarity, was used to evaluate this approach. The evaluation of the BERTCaps model achieved an accuracy of 0.9712 in sentiment classification (binary classification) and 0.8509 in domain classification (multi-class classification).

## 1. Introduction

With the expansion of e-commerce and customer request management systems, a huge volume of textual data is generated directly and indirectly by users every day. Buyers of a product and even managers need to see comprehensive and efficient information that is the result of all opinions about a product or service so that they can make the right decision about the quantity and quality of the product or service provided in the shortest possible time. In other words, the question: "What do people think about x? (x can be a person, subject, feature, product or ...)" is one of the most crucial questions for many people when deciding to buy a product, receive a service or improve them to provide to customers Li, Liu, Zhang, Zhu, Zhu and Yu (2022).

Before the Web became popular, people used to ask their friends and acquaintances a lot of questions about the quality of products and useful recommendations about them when buying a new product. However, this approach only sometimes met their needs because their communication network needed more of the complete experience to respond to all needs. In addition, if an organization or company needed to know the feelings, opinions, and public feedback regarding its constructions and products, it needed to conduct theoretical and group surveys, which, in addition to being limited, also imposed a lot of time and cost Liu (2022).

Most of the information on the Web is in textual and unstructured form. This information does not have any structure or rules. However, reading all this information and concluding about its good or bad takes time and effort. In addition, discovering an inference (positive or negative) when there are several conflicting opinions is difficult even for humans. For this reason, a set of methods called "sentiment analysis" or "opinion mining" was proposed as a subcategory of natural language processing techniques or NLP Liu (2022). In other words, sentiment analysis is a powerful tool that can automatically extract opinions and feelings from online sources with the help of machine learning-based approaches and techniques in natural language processing and classify them as positive, negative or neutral (or in a discrete numerical range, for example, from 0 to 5) Rajput (2020).

The goal of multi-domain sentiment analysis is to train a classifier on a set of labelled data to reduce the need for a large amount of data in specific domains and overcome the challenge of data scarcity in them with the help of data in other domains Yue, Cao, Xu and Dong (2021). In addition, another important issue in multi-domain sentiment analysis is that in order to achieve an accurate classifier, the nature of the training data and the test data must be similar. Therefore, training a model using data from a specific domain and testing it with data from other domains leads to poor classification results Dragoni and Petrucci (2017).

There are various definitions in the literature for sentiment analysis. For example, authors in Pang, Lee et al. (2008) Agarwal, Xie, Vovsha, Rambow and Passonneau (2011) Whitelaw, Garg and Argamon (2005) consider sentiment classifiers to classify unlabeled texts into positive, negative, or neutral categories. Also, authors have considered sentiment analysis (SA) or opinion mining as the computational study of people's opinions, attitudes and feelings toward a topic Pradhan, Vala and Balani (2016) Dragoni and Petrucci (2018). Some other authors have stated the sentiment analysis problem as a 5-set as follows Dragoni and Petrucci (2018):

$$\langle O_j, f_{jk}, so_{ijkl}, h_i, t_l \rangle \tag{1}$$

Where $O_j$ is the target object, $f_{jk}$ is the feature of object $O_j$, and $so_{ijkl}$ is the polarity of the opinion (positive, negative, or neutral) expressed by the opinion holder $h_i$ on the feature $f_{jk}$ at time $t_l$.

ORCID(s):

Multi-domain sentiment analysis is the process of determining the orientation of a text as positive, negative, or neutral using information available from other domains Dragoni and Petrucci (2018). In sentiment analysis, this task is also called text polarity determination. One of this task's most important challenges is determining the polarity of a text that has yet to be used in building a model. Most of the proposed approaches need to improve in classifying such texts. Also known as cross-domain learning, domain adaptation uses labelled data from the source domain (the domain used to train the model) or some target domains (the domain used for testing) to learn a model. The ease or complexity of domain adaptation depends on how closely related the source and target domains are Jeyapriya and Selvi (2015).

Digikala is one of the most important online markets in Iranian internet markets. This is because most people refer to this service to buy items or obtain information about the price or features of products. Thousands of comments about products are recorded there daily, and general sentiments need to be extracted from these comments. On the other hand, in this market, due to the existence of different domains and the lack of comments in each domain, it is necessary to use multi-domain-based models. In Persian, more research needs to be done on analysis, especially multi-domain sentiment analysis. In this regard, in this research, we will examine and present intelligent approaches in Persian for multi-domain sentiment analysis. Given the need for helpful text processing tools such as post taggers and spell checkers and the lack of pre-trained vector representations in Persian, working on this data is more difficult.

## 2. Related Works

Sentiment analysis has been studied in various application domains Wankhade, Rao and Kulkarni (2022)Basiri, Nemati, Abdar, Cambria and Acharya (2021). In this section, we review some common techniques for solving the sentiment analysis problem, which mainly includes traditional machine learning methods and deep learning-based approaches. This issue has also been addressed from both single-domain and multi-domain aspects.

### 2.1. single domain SA

In this analysis, a specific domain is used in sentiment analysis and the model is evaluated in that domain.

### 2.2. Traditional machine learning approaches

- **Supervised learning (SL) approach to sentiment analysis:** SL methods require labelled data. These algorithms require a supervisor to provide the input data and expected outputs Kaur, Mangat et al. (2017). Various supervised learning techniques have been used for sentiment classification. Techniques such as Support Vector Machine (SVM), Naïve Bayes (NB) and Maximum Entropy (ME) have achieved great success in this field Long, Lu, Xiang, Li, Huang et al. (2017).

For example, the authors in Shepelenko (2017) evaluated the performance of three classifiers, NB, ME and SVM, for sentiment analysis on movie reviews. The success of these classifiers is highly dependent on the extracted features. For this purpose, they used unigram, bigram, a combination of unigram and bigram, and a combination of unigram and bigram with part-of-speech labels. The results showed that the NB classifier performs much better on a few features and uses unigram than the other two classifiers. However, when the feature space increases, the SVM classifier achieves better results than NB. Also, in Liu and Lee (2015), NB, SVM and character-based N-gram models were used for sentiment classification on travel and tourism texts and different combinations of machine learning algorithms were investigated using supervised data. Also, in Wahyudin, Djatna and Kusuma (2016), the authors used different classifiers, such as SVM, NB, ME, Random Forest, etc., to classify the sentiment of tweets. The augmented ideas obtained excellent results for classifying the sentiment of tweets. Table 1 is a sample of the literature work using SL approaches for SA.

- **Unsupervised approaches for sentiment analysis:** Supervised approaches require labelled data, which is expensive and difficult to collect. Unsupervised approaches use unlabeled data to discover similar patterns in the inputs. These approaches are used when it is challenging to collect labelled data and easy to collect unlabeled data Van Quang, Chun and Tokuyama (2019). Unlike supervised methods, fewer studies have been conducted on these approaches, of which we will discuss a few in the following. As described by Turney and Littman (2002), this work involves an unsupervised approach to classify comments in either recommendatory or non-recommended. The algorithm that he presented had two basic steps and in the first step, adjectives and adverbs were extracted. In the next step, the conceptual orientation for the phrases extracted in the first phase was calculated. For this, he proposed the PMI-IR algorithm. This algorithm calculates the relations between words using mutual information principles. The following formula demonstrates the calculation of the PMI between two words:

$$PMI(word_1, word_2) = \log_2 \frac{p(word_1, word_2)}{p(word_1)p(word_2)} \quad (2)$$

The algorithm's performance was evaluated on 12 different domains and reached an average accuracy of 74.39%.

In [38], the authors proposed an unsupervised approach to Twitter sentiment classification. They used Parse dependency, a pre-built sentiment dictionary, and linguistic content to categorize tweets. Xiao,

| Approaches | References |
|---|---|
| Support Vector Machine (SVM) | Mullen and Collier (2004)Gupta, Tyagi, Choudhury and Shamoon (2019)Gupta et al. (2019) |
| Decision Tree (DT) | Singh and Tripathi (2021)Neelakandan and Paulraj (2020) |
| Random Forest (FR) | Al Amrani, Lazaar and El Kadiri (2018) |
| Naïve Bayes (NB) | Abbas, Memon, Jamali, Memon and Ahmed (2019) |
| Logistic Regression (LR) | Al Omari, Al-Hajj, Hammami and Sabra (2019) |
| Conditional Random Field (CRF) | Wang, Pan, Dahlmeier and Xiao (2016) |
| Hidden Markov Models (HMMs) | Perikos, Kardakis and Hatzilygeroudis (2021) |

**Table 1**
Traditional Machine learning(ML) approaches for sentiment analysis.

| Approaches | References |
|---|---|
| Convolution Neural Networks (CNNs) | Flandhi, Malarvizhi Kumar, Chandra Babu and Karthick (2021)Usama, Ahmad, Song, Hossain, Alrashoud and Muhammad (2020) |
| Long Short Term Memory (LSTM) | Shobana and Murali (2021)Muhammad, Kusumaningrum and Wibowo (2021) |
| Capsule Network | Obied, Solyman, Ullah, Fat'hAlalim and Alsayed (2021)Elhorbanali and Sohrabi (2024) |
| flated Recurrent Unit (flRU) | Xu, Chen, Ding and Wang (2024) Mohammad, Hammad, Sa'ad, Saja and Cambria (2023) |
| Transformer | Tan, Lee, Anbananthen and Lim (2022) Kokab, Asghar and Naz (2022) |

**Table 2**
Deep learning approaches for sentiment analysis.

Zhang, Chen, Wang and Jin (2018) proposed an unsupervised approach to sentiment analysis in Chinese using a set of automatically selected base words. Their main goal was to improve the accuracy of their classifier by automatically selecting base words. Two assumptions were made for the selection of these words. The first assumption is that candidate words selected for the base may be associated with negative words. For example, it is widespread in Chinese to use the word 'Not good' instead of 'bad'. The second assumption is that the polarity of the base words should be specified. To do this, they used the word 'good' as a standard for positive words and compared the patterns in which the base words are found with the patterns shown in the standard (gold) dataset. The results showed that their approach was close to supervised approaches and, in some cases, even better than them.

## 2.3. Deep learning-based approaches

Deep learning includes various types of neural networks such as ANN, CNN, RNN, LSTM, GRU, and CapsuleNet. Below, we will briefly discuss some of the work done with these networks. Table 2 is a classification of the work done in the literature based on these networks to solve the problem of sentiment analysis:

In multi-domain sentiment analysis, as mentioned, the goal is to train a classifier on a set of domains, so that the problem of domain dependency is solved. In Dragoni and Petrucci (2017), an embedded representation of words was studied along with a deep learning architecture for performing multi-domain classification. The main idea of the presented approach is to transform the input raw texts into an embedded representation based on word2vec and use it to predict the polarity of the comments, and to identify the domain of belonging in parallel with the polarity discovery operation. In this approach, LSTM were used to create a deep network. In this paper, the output value of the beam layer was used to determine the domain of belonging as follows:

$$y = softmax(W_y c + b_y) \qquad (3)$$

The beam layer is also used to determine the polarity of each domain as follows:

$$Z = tanh(W_z c + b_z) \qquad (4)$$

In Petrucci and Dragoni (2016) and Dragoni and Petrucci (2018), two similar approaches were presented to solve this problem. Their general goal was to exploit domain-specific information to build models that combined this information to achieve a general overlap. They also used two linguistic dictionaries, SentiNet and General Inquirer, to compensate for the amount of information lost in combining different domains.

Most of the studies conducted in SA have been conducted on natural languages such as English, Chinese and Arabic. NLP in Persian and Arabic is still in its early stages Farghaly and Shaalan (2009). It lacks advanced resources and tools. Therefore, Persian and Arabic still faces challenges in NLP tasks due to its complex structure, history and different cultures Rushdi-Saleh, Martín-Valdivia, Ureña-López and Perea-Ortega (2011). A large number of tools and approaches, which are either semantic approaches or machine learning (ML) approaches, have been used in the literature to perform SA tasks. Most of them are designed to handle SA in English, which is a scientific language. The semantic approach extracts emotional words and calculates their polarities based on emotional words. In contrast, to build a new model, ML classifiers are trained on the annotated data after converting them into feature vectors to infer specific features used in a particular class. Finally, the new model can be used to predict the class of new data. It is worth noting that these approaches can be adapted to other languages such as Arabic. Arabic has received less effort compared to other languages; however, hundreds of studies have been proposed for ASA. Since its introduction over a decade ago, ASA has become one of the most popular forms of extracting information from surveys. Table 3 summarizes examples of these approaches for Arabic and Persian.

## 3. Methodology

CNNs work by collecting features at each layer. Deeper layers (closer to the input) learn more superficial features.

| Model | Reference | Domain Type | Model type | Performance |
|---|---|---|---|---|
| flRU | Abdelgwad, Soliman, Taloba and Farghaly (2022) | Single Domain | Deep learning | Accuracy= 83.98% |
| SVM and Naive Bayes | Mohamed (2022) | Single Domain | Machine learning | SVM Accuracy=91.40% Naive bayes Accuracy=88.08% |
| TFICF | Alqmase and Al-Muhtaseb (2022) | Multi-Domain | Lexicon Based | Accuracy= 89% |
| Arabic Ontology-Based | Khabour, Al-Radaideh and Mustafa (2022) | Multi-Domain | Ontology-Based | Accuracy= 79.20% |
| Different classification and different feature selection approaches | Habib (2021) | Single domain | - | Precision=91.22% Recall=91.71% F1=91.46% |
| Unsupervised models and neural networks | Akhoundzade and Devin (2019) | Single domain | CNN | Precision=73.7% Recall=99.1% F1=58.6% |
| Conceptual dictionary of words and polarity recognition | Asgarian, Kahani and Sharifi (2018) | Single domain | - | Accuracy=86% F1=80% Recall =75% |

**Table 3**
Single-domain and multi-domain sentiment analysis on Arabic and Persian data.

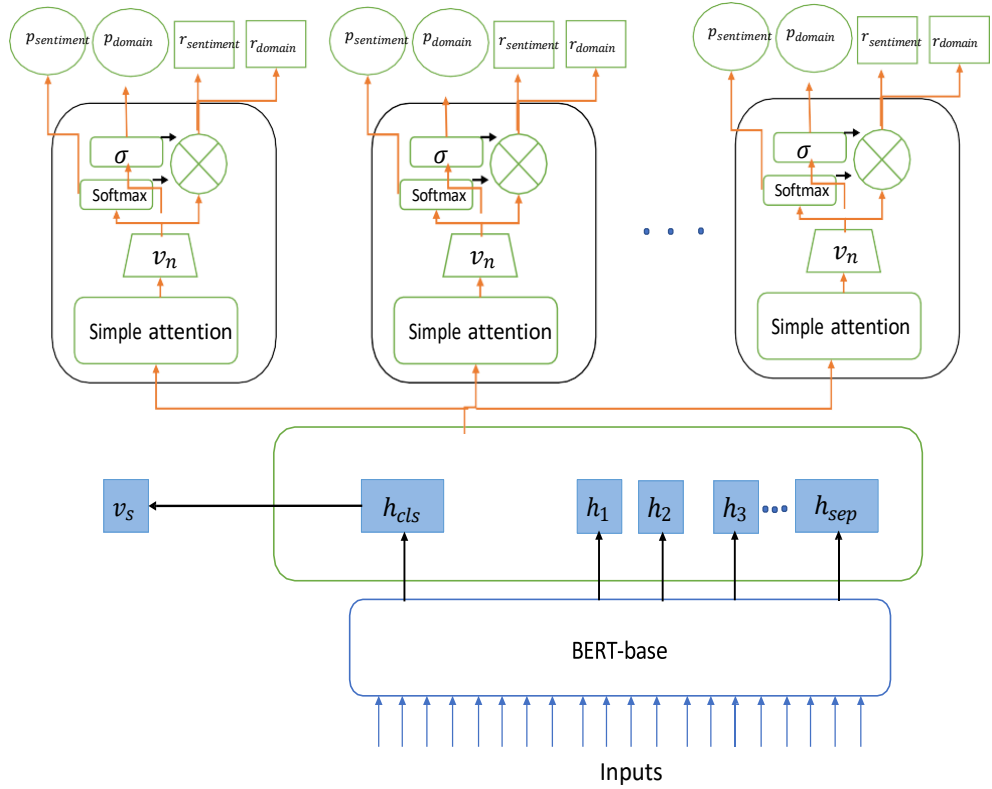

**Figure 1:** BERT-Capsule model architecture.

Higher layers combine simple features with complex features, finally leading to classification. For example, in an image, they start by finding edges, shapes, and real objects and learning edges and colour changes (simple features) in deep layers. Similarly, the same thing happens when using CNN on textual data. In other words, regardless of the method of generating input vectors from Word2vec text (or Glove or any other method), simple features in deep layers and complex features in higher layers are learned and lead to classification. In a CNN network, layer pooling reduces the amount of calculations and makes the network less dependent on the input features. which part of the input is concerned. So, the input data may also be given to the network. When transformed data is given to the network to extract its features, it may be in the extraction features may be wrong and thus lead to wrong classification, so it is necessary to give different data as input to the network so that the network does not make mistakes (for example, different rotations of the same image) [60].

Capsule networks [60] have been presented to solve these problems. Capsule means cover and protection, and it is used because all the essential features of the input data are preserved by this method. The important difference between capsule networks and convolution is that, unlike convolution, where values are stored numerically (scalar), they are stored in the capsule in vector form. Every vector, in its nature, has two characteristics: size and direction. In the capsule, the presence of a specific feature is modelled by the size of the vector and its change by the direction of the vector, so by changing the parameters of a specific feature, the size of the vector (the feature itself) is fixed but its direction (parameters) changes. In capsule networks, complex calculations are performed on features, and the results of these calculations are mapped to a vector containing beneficial information. Therefore, each capsule learns to observe and change the appearance of entities (features) in a limited range of conditions and two possibilities: one is the presence of features in a limited range, and the other is a set of exploratory parameters related to this feature produced as output. If the capsule works correctly, the probability of the main feature itself is unchangeable, while the probability of the heuristic parameters associated with these features is variable [60].

Different approaches have used capsule networks for textual data, including this network for sentiment analysis [61-63], text classification [64-66], etc. Notably, these types of networks require convolutional networks or RNNs to create initial vectors. In the proposed approach, we use BERT encoder to create initial vectors.

An overview of the proposed approach is shown in Figure 1. In the following, we will discuss the details of this model.

## 3.1. BERT Encoder

In this model, the BERT basic encoder is used. BERT, a pre-trained deep bidirectional transform model, produces context-rich word embeddings that outperform RNN encoders. BERT's critical technical innovation is using two-way Transformer training, a popular attention model, to model language. The BERT results show that a bidirectionally trained language model can have a deeper sense of language texture and flow than unidirectional language models. BERT is an encoder stack of transformer architecture. Transformer architecture is an encoder-decoder network that uses self-attention on the encoder side and attention on the decoder side. BERTBASE has 12 layers in the Encoder stack, while BERTLARGE has 24 layers in the Encoder stack Ramirez, Velastin, Cuellar, Fabregas and Farias (2023). The BERT architectures (BASE and LARGE) also have more extensive feed-forward networks (768 and 1024 hidden units, respectively) and more attention heads (12 and 16, respectively) than the Transformer architecture proposed in the original paper. BERTBASE contains 110 million parameters, while BERTLARGE contains 340 million parameters. This model first takes the CLS token as input and then follows a sequence of words as input. Here, CLS is a classification token. It then sends the input to the upper layers. Each layer applies its attention, passes the result through a feed-forward network, and then delivers it to the next encoder. The model outputs a vector of hidden size (768 for BERT BASE). If we want to get the output of a classifier from this model, we can get the output corresponding to the CLS token.

In this model, the output from BERT's final hidden layer is considered the input tensor $H$ for the capsule network. In fact, by removing the role of convolutional and recurrent layers, BERT can provide richer vectors and effectively depict its emotional differences.

## 3.2. Instance Representation

Instance representation refers to a vector representation that includes general semantic information of a sentence or document Guo (2024). In RNN-based models, where the input sequence is dependent, this representation is the average of all word vectors in a sentence. Unlike classical deep learning and feedback learning models, BERT is inherently a complete sentence representation vector that provides the final output hidden vector $h_{[CLS]}$. As mentioned, the output of the CLS token can be used for classification. To represent the sample $v_s$, the hidden layer vector corresponding to the token $[CLS]$ from the last BERT layer is used. The $[CLS]$ token in BERT is specifically designed for classification tasks, with its hidden state vector trained to sum up the representation of the entire input sample, making it very suitable for representing the sentiment of a text sample.

## 3.3. Capsule Structure

The structure of the proposed capsule includes three basic modules. These modules are discussed in three separate subsections.

### 3.3.1. Representation Module

In this module, the outputs of the attention mechanism are used to create the display of capsules. By summing the outputs of the hidden layer of the BERT encoder, each capsule obtains a vector representation corresponding to a category of emotions [6]. This module is a simple attention layer whose input is the output H of the final hidden layer of the BERT base: $H = [h_{[CLS]}, h_1, h_2, \ldots, h_n, h_{[SEP]}]$. Let $w_i$ be the weight vector for the $i$-th capsule. With these interpretations, the attention score $\alpha_i$ is calculated as follows:

$$\alpha_i = \text{softmax}(H \cdot w_i) \tag{5}$$

$v_i$ is the representation vector and calculated as:

$$v_i = \alpha_i^T \cdot H \tag{6}$$

With these interpretations, it can be shown:

$$v_i = \text{softmax}(H \cdot w_i)^T \cdot H \tag{7}$$

$v_i$ is an index that is considered as an input for the representation module and the probability module.

### 3.3.2. Probability Module

This module calculates the activation probability for each capsule. According to the form of the research problem (sense classification; domain classification), the activation probability pair is used for this layer. This module reflects the match of the sentiment category and the domain it represents with the sentiment of the input instance and the domain of the input instance. The capsule with the highest activation probability is considered to be the prediction of the input sample's sentiment and the input domain's prediction. The probability module is a fully connected layer that maps the representation vector $v_i$ to the capsule activation probability:

$$p_{sentiment} = \sigma(u_i \cdot v_i + b_i) \tag{8}$$

$$p_{domain} = Softmax(u_i \cdot v_i + b_i) \tag{9}$$

### 3.3.3. Reconstruction Module

The purpose of this module is to display the emotions of the input sample based on the capsule display activated in the previous step. By minimizing the reconstruction error between the reconstructed emotion representation and the original emotion representation and the reconstruction error between the reconstructed domain representation and the original domain representation, we can increase the prediction accuracy of the model. The Reconstruction module multiplies the activation probability of the capsule $p_i$ by the representation vector of the capsule $v_i$, which results in the reconstructed representation vector $r_i$.

$$r_{sentiment} = p_{sentiment}.v_i \qquad (10)$$

$$r_{domain} = p_{domain}.v_i \qquad (11)$$

## 4. Data

The lack of a multi-domain protocol in Persian has presented serious challenges to multi-domain sentiment analysis. We used the Digikala[1] dataset to evaluate the proposed model. In this regard, 50,799 comments were collected in 10 domains: shoes, perfume, phone, cream, printer, clothes, books, beds, cars, and gold. Python and the Beautiful Soup Library were used to collect the data. An attempt was also made to normalize the texts before labelling, for which Hazem was used. The data preprocessing process was applied before labelling, such as removing empty comments, removing URLs, removing stickers, and removing numbers. The labelling process was completely manual. In fact, after thoroughly reviewing the comments, the collection team adopted two positive and negative labels for this data. Also, in choosing the polarity of the comments, comments with a score of 4 or higher were considered positive, and those with a score of less than 2 were considered negative. It is worth noting that the collected data could be more balanced, and the number of negative samples is much lower than the number of positive samples. Figure 2 shows the percentage of data frequency per label, data frequency per domain, and label per domain.

## 5. Result

At first, we divided the Digikala dataset into training and testing data. For this purpose, 80% of the data was used for training and 20% for testing. The Keras[2] library was used to develop the proposed models. Keras is a high-level library written in Python. This API can run seamlessly in both GPU and CPU environments. The hardware and software specifications used in this research are listed in Table 4. Criteria such as accuracy, precision, recall, and F1 score were also used to evaluate the proposed model. These criteria can be defined according to the following relationships:

---
[1]https://www.digikala.com/ [2]keras.io

$$Accuracy = \frac{(TP + TN)}{(TP + TN + FP + FN)} \qquad (12)$$

$$Precision = \frac{TP}{TP + FP} \qquad (13)$$

$$Recall = \frac{TP}{TP + FN} \qquad (14)$$

$$F1 - score = 2 * \frac{precision * recall}{precision + recall} \qquad (15)$$

We examined different basic models for this purpose. In the following, we will give a brief explanation of each of the models.

- **CNN-Multi Channel**: In Zhang and Wallace (2015), a simple CNN network with a convolution layer on top of word vectors derived from a non-supervised language model is taught. This architecture is both dynamic and static, which determines the model's training on embedded vectors. This architecture was used as another architecture to compare with the proposed approaches.

- **Character level CNN**: In this case, the modelled input consists of a string of encrypted characters. The encryption process involves choosing an m-sized alphabet for the target language and transforming each character into a vector representation using m-to-1 encoding (aka "one-hot" encoding). In effect, the string of characters turns into a string of such m-length vectors. The alphabet used in all models comprises 75 characters, 32 of which are Persian alphabets, 10 are numeric digits, and 33 are punctuation marks. This model was employed to provide a comparison with the proposed models. Details of this model's architecture can be acquired from Zhang, Zhao and LeCun (2015).

- **NeuroSent**: The essence of this strategy Dragoni and Petrucci (2017) lies in transforming the textual inputs into an embedded representation to assess the sentiment of given comments. In addition, determining the domains is performed simultaneously with the exercise of predicting its direction of the sentiment. Within this concept, the skip-gram Word2vec model is used to acquire the constructions of words. At the center of this concept of building a deep network, is the use of LSTM memory cells. This architecture serves the purpose of learning the general characteristics of the domains and has utilized these characteristics to train the network.

- **Bi-GRUCapsule**: This model has been investigated in Khayi and Rus (2019). The performance of this

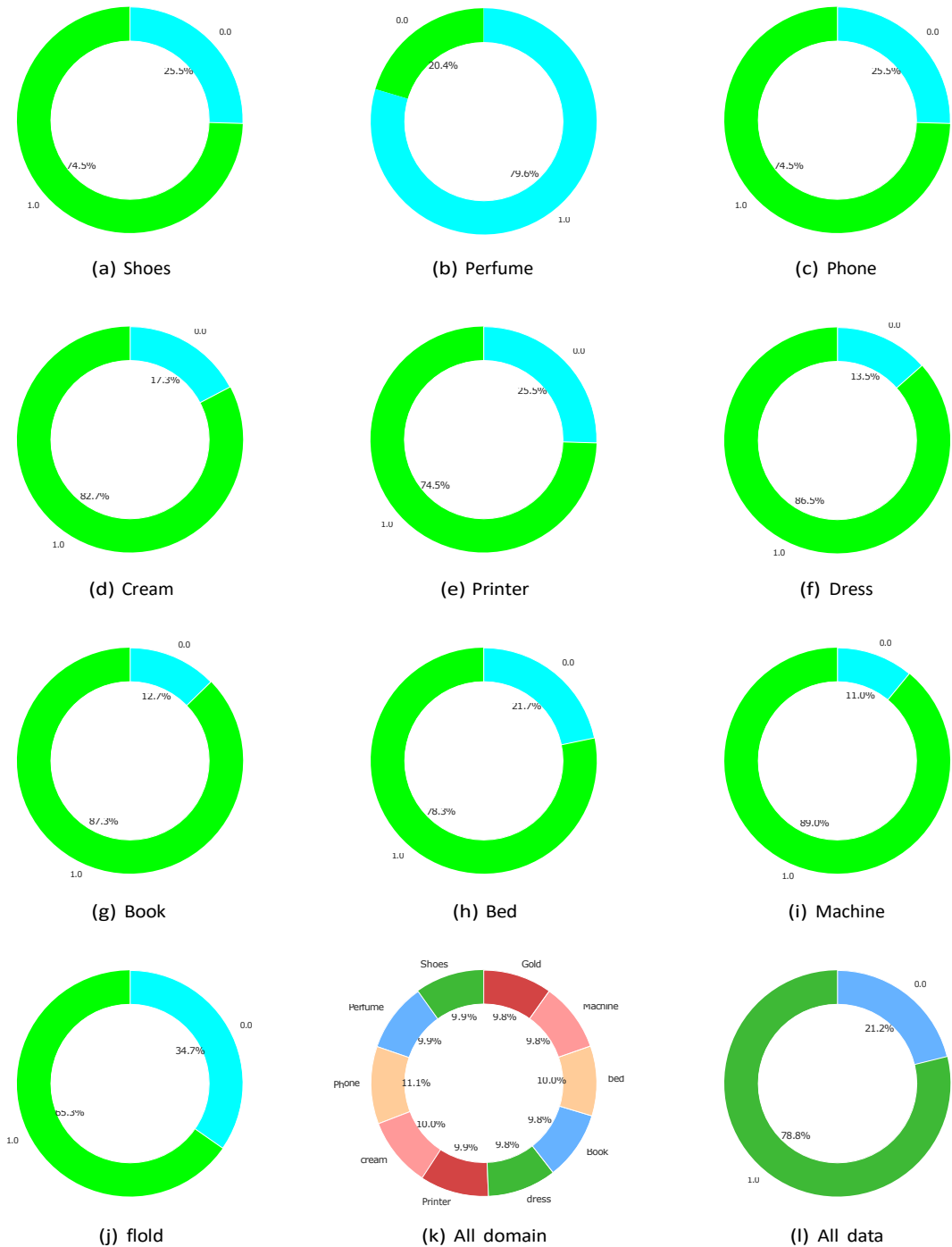

**Figure 2:** Statistical analysis of the research data set, which includes the analysis of each domain alone and all domains together.

model in previous textual and image data has been investigated in [68]. In this model, Bi-GRU is used to extract initial vectors, and Capsule is used to learn the extracted features.

- **INDCAPS:** The INDCAPS approach combines two bidirectional networks of IndRNN and CapsuleNet to solve the multi-domain SA problem, where Bi-GRU has the role of extracting features for CapsuleNet, and

CapsuleNet in this model is responsible for learning features Mousa, Dadgostarnia, Olfati Malamiri, Behnam and Mohammadi (2024).

Table 5 shows the results of polarity detection. For this purpose, the sigmoid function was used in the last layer of each network. The CNN-Multi Channel model achieved an accuracy of 0.8704 in the training data. This approach

| Software | | |
|---|---|---|
| Name | version | Description |
| Ubuntu Bionic Beaver (LTS) | 18.04.2 | Operating System |
| Python | 3.6.7 | Used for implementation |
| Keras | 2.2.4 | Used for building models |
| Pandas | 0.23.4 | Used for data analysis |
| Tensorflow | 2.12.0 | Used as backend for Keras |
| CUDA | 9.0.176 | Required for Tensorflow |
| cuDNN | 7.4.1 | Required for Tensorflow |
| Hardware | | |
| Name | Version | |
| CPU | Intel i7-2600 | |
| GPU NVIDIA | GeForce GTX 980 | |
| Memory | Kingston 8 GB DDR3 | |
| GPU Memory | 4 GB, GDDR5 | |

**Table 4**
The hardware and software specifications of this research.

| Model | Polarity detection | | | | | |
|---|---|---|---|---|---|---|
| | Accuracy | | Precision | | Recall | |
| | Train | Test | Train | Test | Train | Test |
| CNN-Multi Channel | 0.8704 | 0.8775 | 0.8834 | 0.8627 | 0.9331 | 0.9207 |
| Character level CNN | 0.9002 | 0.8823 | 0.9220 | 0.9212 | 0.9312 | 0.9178 |
| NeuroSent | 0.9160 | 0.9183 | 0.9290 | 0.9234 | 0.9222 | 0.9118 |
| Bi-GRUCapsule | 0.9423 | 0.9345 | 0.9231 | 0.9347 | 0.9432 | 0.9336 |
| Bi-IndRNNCapsule | 0.9565 | 0.9489 | 0.9706 | 0.9535 | 0.9744 | 0.9524 |
| BERTCaps | **0.9712** | **0.9576** | **0.9855** | **0.9623** | **0.9881** | **0.9671** |

**Table 5**
The results obtained by BERTCaps and different approaches on training and test data for Polarity detection reached an accuracy of 0.8775 in the test data. The character-level CNN model achieved an accuracy of 0.8823 in the test data set. The value of precision and recall in this approach was equal to 0.9212 and 0.9178, respectively. The NeuroSent approach was able to achieve an accuracy of 0.9183 in test data and an accuracy of 0.9160 in training data. Bi-GRUCapsule and Bi-IndRNNCapsule, which are completely similar in terms of structure, with the difference that they use two Bi-GRU and Bi-IndRNN gates, were able to achieve 0.9345 and 0.9489 accuracy in the test data, respectively. The proposed BERTCaps approach achieved 0.9712 accuracy in training data and 0.9576 accuracy in test data. This approach also achieved precision and recall of 0.9623 and 0.9671, respectively, in the test data. The remarkable thing about sentiment classification is that capsule-based approaches performed more successfully.

Table 6 also shows the results obtained by different approaches in domain identification. In this task, CNN-Multi Channel achieved an accuracy of 0.6958. This approach obtained the weakest result among the tested approaches. Meanwhile, the Character level CNN approach achieved 0.7043 accuracy. This approach also achieved precision=0.8653. The NeuroSent approach achieved an accuracy of 0.7377 in the 10 domains tested. Capsule-based approaches were also able to achieve high accuracies in this task. The Bi-GRUCapsule approach achieved an accuracy of 0.7809, and the Bi-IndRNNCapsule approach achieved an accuracy of 0.8020. In the meantime, the proposed BERTCaps approach reached an accuracy of 0.8509. This approach also achieves precision=0.9432 and recall=0.9112.

Table 7 shows the results obtained by these approaches for domain identification. BERTCaps approach has achieved higher accuracy than other approaches. Accuracy is generally due to different domains and linguistic differences between domains. The NeuroSent approach achieved a median accuracy of 0.8861 in domain evaluation. This approach achieves better results than the three models: Multi-channel CNN, Character level CNN, and Bi-GRUCapsule. The Multi-channel CNN approach obtained better results than all the tested approaches. In this approach, the worst result was obtained in the printer domain. In Character-level CNN, the printer and phone approach obtained the worst result. Capsule-based approaches such as Bi-GRUCapsule, Bi-IndRNNCapsule, and BERTCaps achieved an average accuracy higher than 0.8800. The success of these models is due to the use of capsule vectors. In the Bi-GRUCapsule model, the highest accuracy is obtained in the printer range. In Bi-IndRNNCapsule, the average accuracy is equal to 0.9144. This approach reached an accuracy of 0.9291 in the domain of cold cream, which was the highest accuracy. BERTCaps achieved an average accuracy of 0.9277. The highest and lowest accuracy obtained by this approach is for the domain of bed and jewelry, respectively.

| Model | Domain identification | | |
|---|---|---|---|
| | Accuracy | Precision | Recall |
| CNN-Multi Channel | 0.6958 | 0.8582 | 0.7255 |
| Character level CNN | 0.7043 | 0.8653 | 0.7589 |
| NeuroSent | 0.7377 | 0.8982 | 0.8290 |
| Bi-GRUCapsule | 0.7809 | 0.8909 | 0.8554 |
| Bi-IndRNNCapsule | 0.8020 | 0.9223 | 0.8829 |
| BERTCaps | **0.8509** | **0.9432** | **0.9112** |

**Table 6**
The results obtained by BERTCaps and different approaches on test data for Domain identification.

| Domain | NeuroSent | Multi-channel CNN | Character level CNN | Bi-GRUCapsule | Bi-IndRNNCapsule | BERTCaps |
|---|---|---|---|---|---|---|
| Shoes | 0.8717 | 0.8232 | 0.8514 | 0.8692 | 0.9134 | **0.9246** |
| perfume | 0.8837 | 0.8122 | 0.8420 | 0.8864 | 0.8917 | **0.9024** |
| phone | 0.8718 | 0.8105 | 0.8097 | 0.8693 | 0.9232 | **0.9434** |
| cold cream | 0.8650 | 0.8402 | 0.8728 | 0.8652 | 0.9291 | **0.9323** |
| printer | 0.8866 | 0.8054 | 0.8093 | 0.8915 | 0.9272 | **0.9567** |
| dress | 0.8996 | 0.8434 | 0.8714 | 0.8762 | **0.9075** | 0.8909 |
| Book | 0.8941 | 0.8455 | 0.8700 | 0.8926 | 0.9094 | **0.9232** |
| Bed | 0.8911 | 0.8612 | 0.8543 | 0.8911 | 0.9114 | **0.9139** |
| Shaving machine | 0.9086 | 0.8491 | 0.8704 | 0.8806 | 0.9125 | **0.9324** |
| jewelry | 0.8890 | 0.8567 | 0.8589 | 0.8828 | 0.9193 | **0.9573** |

**Table 7**
Detailed Results Obtained on Domains Contained within the Digikala Dataset by the Baselines and by BERTCaps.

The results of different folds in k-fold cross validation were tested by different approaches and the proposed approach on domain identification is shown in Figure 3-a.

The highest result for CNN-Multi Channel was obtained in fold 1, also mean=0.6831 and std=0.012 values. Mean and std values for Character level CNN are equal to 0.6918 and 0.0110 respectively. In the NeuroSent approach, the highest accuracy is achieved in fold 4. The std values of the three approaches GRUCapsule, Bi-IndRNNCapsule and BERTCaps are 0.0095, 0.0063 and 0.0014 respectively. Also, the t-test value of the proposed approaches in K-field evaluation is shown in Table 8.

The k-fold results of different approaches in polarity detection are shown in Figure 3-b, and the t-test evaluation of these approaches is shown in Table 9. The highest result obtained in CNN-Multi result was obtained in fold2. Also, compared to most of the approaches, the highest result was obtained in fold 1. In the proposed approach, the highest result was obtained in fold 5.

### 5.1. Discussion

Various approaches have been proposed in the literature to analyze multi-domain and single-domain Persian SA. Most of these approaches have been developed with a focus on a specific domain. These approaches have been investigated in the form of machine learning and deep learning. In this research, a model based on Brett and Capsule was presented for Persian multi-domain classification. The purpose of the proposed approach was to focus on 10 different areas. This approach was used for polarity detection and domain identification. BRET was mined for feature representation and CapsuleNet for feature learning. The efficiency of this algorithm was evaluated in comparison with 5 other approaches and it was shown that the proposed approach achieved higher accuracy in both tasks. Figures 4 also show the number of TP, FP, FN and TN rates in different domains. As shown in Figure 26-17, the Shoes and Perfume domain has the highest FP value and the Perfume and dress domain has the highest TN value. These results were also deduced by the Bi-IndRNN+Capsule approach.

## 6. Conclusion

Sentiment classification is a primary task in natural language processing, assigning one of the three positive, negative or neutral classes to free texts. However, as shown by many researchers, sentiment classification models are highly domain-dependent. The classifier may perform classification with good accuracy in one domain but not in another due to the semantic multiplicity of words' poor accuracy. The purpose of multi-domain sentiment analysis is to train a classifier based on a set of labelled data to reduce the need for a large amount of data in specific domains and overcome the challenges of data scarcity in them with the help of fixed data in other domains. The purpose of this article was to present a new method for Persian multi-domain sentiment analysis using capsule networks and the BERT approach. BERT was used for feature representation, and Capsule was used for feature learning. The Digikala data set was used to evaluate the proposed model. Ten different domains were considered for this purpose. The proposed method was evaluated using the Digikala dataset, and acceptable accuracy was obtained compared to the existing approaches. It achieved an average deft of 0.9277 on different domains in k-fold evaluation. One of the remaining challenges in the proposed approach is handling the unbalanced class. Data augmentation techniques can be used for this purpose. Also, using class-sensitive functions can be helpful in handling this challenge. Another remaining challenge in the proposed approach is the parameter space. For this purpose, PSO, GWO, and IGWO should be used in future works.

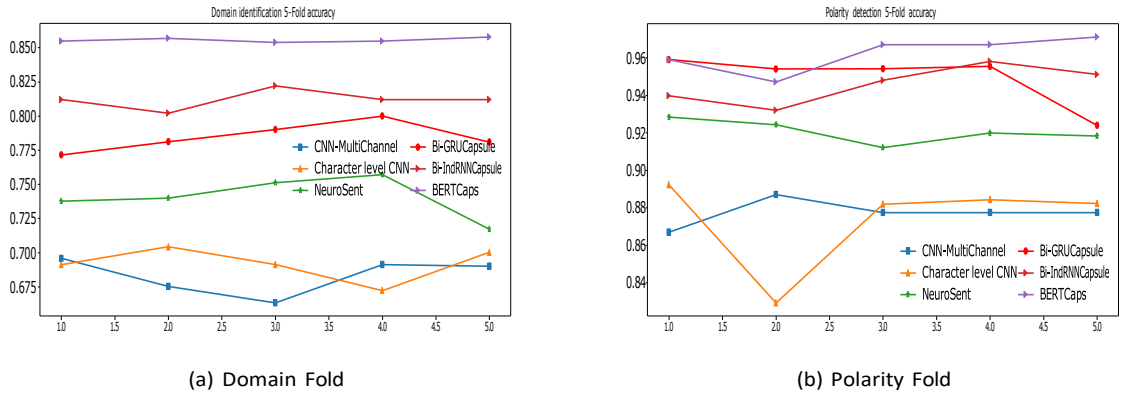

**Figure 3:** Domain identification and Polarity detection k-fold cross validation.

| Model | CNN-Multi Channel | Character level CNN | NeuroSent | Bi-GRUCapsule | Bi-IndRNNCapsule | BERTCaps |
|---|---|---|---|---|---|---|
| CNN-MultiChannel | 1.0 | 0.3151 | 0.0002 | 1.0355 | 6.3205 | 2.5492 |
| Character level CNN | 0.3151 | 1.0 | 0.00055 | 1.3830 | 6.4427 | 1.9306 |
| NeuroSent | 0.0002 | 0.0005 | 1.0 | 0.0007 | 1.3128 | 1.7171 |
| Bi-GRUCapsule | 1.0355 | 1.3830 | 0.0007 | 1.0 | 0.0014 | 4.6293 |
| Bi-IndRNNCapsule | 6.3205 | 6.4427 | 1.3128 | 0.0014 | 1.0 | 9.2914 |
| BERTCaps | 2.5492 | 1.9306 | 1.7171 | 4.6293 | 9.2914 | 1.0 |

**Table 8**
T-test values(P-values) obtained by different testing approaches(Domain identification).

| Model | CNN-Multi Channel | Character level CNN | NeuroSent | Bi-GRUCapsule | Bi-IndRNNCapsule | BERTCaps |
|---|---|---|---|---|---|---|
| CNN-Multi Channel | 1.0 | 0.7890 | $0.0006e-6$ | $0.0007e-6$ | $0.0001e-6$ | $0.0002e-6$ |
| Character level CNN | 0.7890 | 1.0 | 0.0040 | 0.00041 | 0.0003 | $0.0008e-6$ |
| NeuroSent | $0.0006e-6$ | 0.0040 | 1.0 | 0.0032 | 0.0014 | $0.0003e-6$ |
| Bi-GRUCapsule | $0.0007e-6$ | 0.0004 | 0.0032 | 1.0 | 0.6631 | 0.1305 |
| Bi-IndRNNCapsule | $0.0001e-6$ | 0.0003 | 0.0014 | 0.6631 | 1.0 | 0.0292 |
| BERTCaps | $0.0002e-6$ | $0.0008e-6$ | $0.0003e-6$ | 0.1305 | 0.0292 | 1.0 |

**Table 9**
T-test values(P-values) obtained by different testing approaches(Polarity detection).

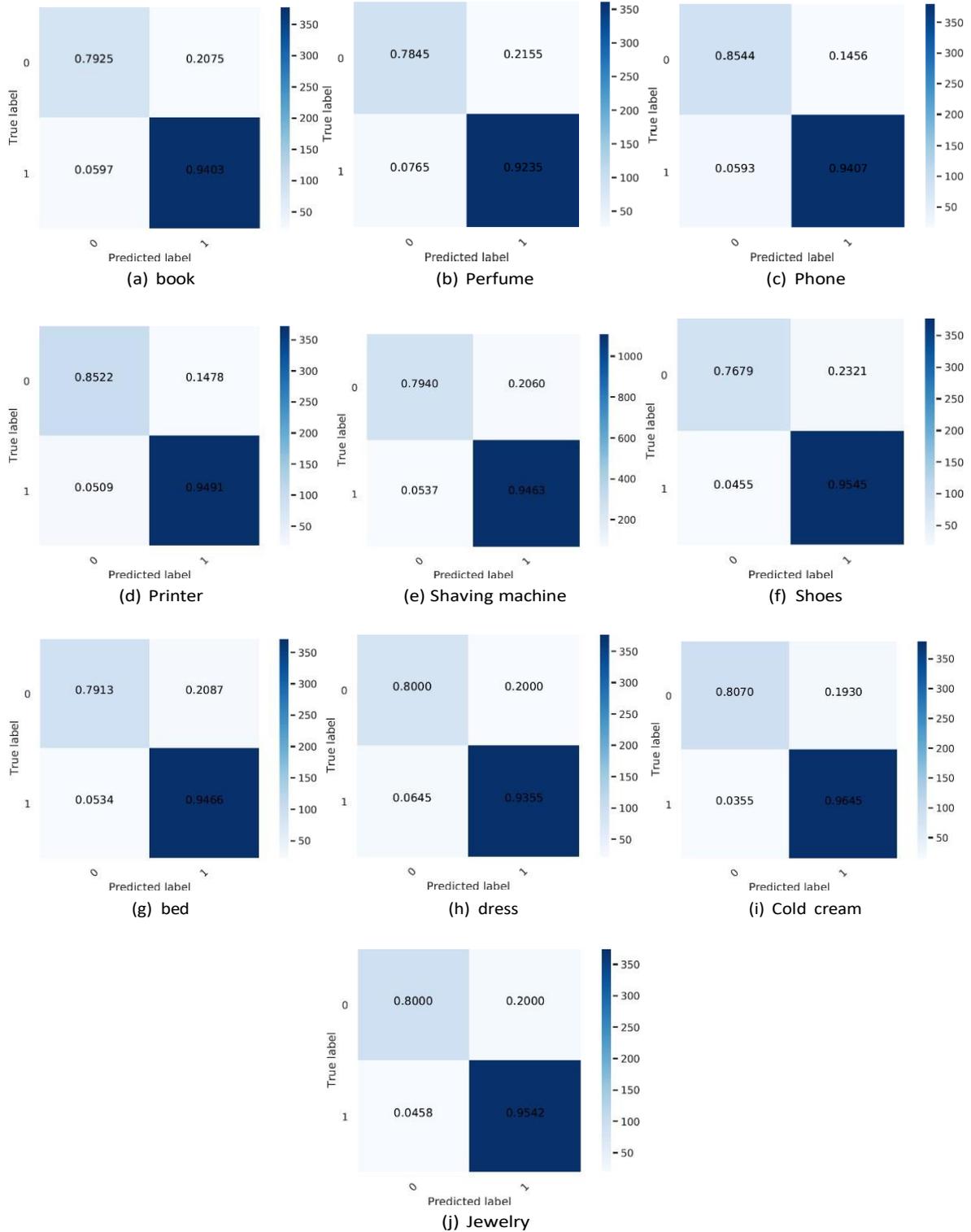

**Figure 4:** Confusion matrices for the predictions on the Digikala dataset.